%% file: ms.tex
\documentclass[letterpaper, 11 pt, conference, onecolumn]{ieeeconf}  % Comment this line out if you need a4paper

\IEEEoverridecommandlockouts                              % This command is only needed if 
\usepackage{graphicx} % for pdf, bitmapped graphics files
\usepackage{mathptmx} % assumes new font selection scheme installed
\usepackage{times} % assumes new font selection scheme installed
\usepackage{amsmath} % assumes amsmath package installed
\usepackage{amssymb}  % assumes amsmath package installed
\usepackage{cleveref}
\usepackage{subcaption}
\usepackage{placeins}
\usepackage{multicol}
\usepackage[]{apacite}
\title{\LARGE \bf
Learning to Roam Free from Small-Space Autonomous Driving with A Path Planner
}

\author{Sascha Hornauer$^{1}$, Karl Zipser$^{2}$ and Stella Yu$^{3}$% <-this % stops a space
\thanks{$^{1}$UC Berkeley
        {\tt\small sascha.hornauer@icsi.berkeley.edu}}%
\thanks{$^{2}$UC Berkeley
        {\tt\small karlzipser@berkeley.edu}}%
\thanks{$^{3}$UC Berkeley
        {\tt\small stellayu@berkeley.edu}}%
}
\begin{document}

% General outline
\maketitle
\thispagestyle{plain}
\pagestyle{plain}

\input{1abs.tex}
\input{2intro.tex}

\input{3related.tex}

\input{4method.tex}

\input{5results.tex}
\input{6summary.tex}
\clearpage
\bibliographystyle{apacite}
\bibliography{planner.bib,mendeley.bib}
\end{document}

%% file: 1abs.tex
\begin{abstract}

 Modern autonomous driving algorithms often rely on learning the mapping from visual inputs to steering actions from human driving data in a variety of scenarios and visual scenes.  The required data collection is not only labor intensive, but such data are often noisy, inconsistent, and inflexible, as there is no differentiation between good and bad drivers, or between different driving intentions.
 We propose a new autonomous driving approach that learns roaming skills from an optimal path planner.  Our model car practices reaching random target locations in a small room with obstacles, by following the optimal trajectory and executing the steering actions decided by a planner.  We learn the associations of driving behaviours with depth images, instead of raw color images of the visual scene.  This more universal spatial representation allows the learned driving skills to transfer immediately to novel environments with different visual appearances.
 Our model car trained in a simple room, void of many visual features, demonstrates surprisingly good driving performance in a cluttered office environment, avoiding collisions with novel obstacles and unseen layouts of drive-able space. Its performance on outdoor curbside driving is also on par with human driving.

\end{abstract}

% \begin{keywords}
% autonomous driving, path planning, collision avoidance, depth image
% \end{keywords}

%% file: 2intro.tex
\section{Introduction}

We consider the task of training a convolutional neural network that allows a model car equipped with a stereo camera and mobile GPU computing power to roam free, a challenge i.e. in collision-free autonomous driving, in unseen and possibly cluttered environments. 
There are a few main approaches. 1) Traditional approaches analyze the sensory input and develop a catalogue of path planning and driving policies under various scenarios \cite{Plessen2018,Hubmann2017,Bittel2017,Kong2015,Seff2016}. Such approaches require a lot of engineering efforts and are often brittle in real applications. 2) Reinforcement learning approaches allow the model car to discover the right driving policy on the fly through trials and errors \cite{Kahn2017,Kuderer2015,Mirowski2016} However, such approaches are often slow and most research is done with simulated and simple environments without much clutters.
3) Data-driven learning approaches that turn the visual input into driving actions directly \cite{Codevilla2017,Pan2017,Chowdhuri2017,Hou2017,Bojarski2016a,LeCun2006,Pomerleau1989}. With the availability of modern big data, computing power, and deep learning techniques, deep-net based end-to-end driving systems become a major research direction. Such approaches, while versatile, require extensive human driving data in a variety of actual environments. They often fail when the visual scene at the test time does not match those during training. 
\begin{figure*}[!htbp]
    \centering
    \includegraphics[width=1.0\textwidth]{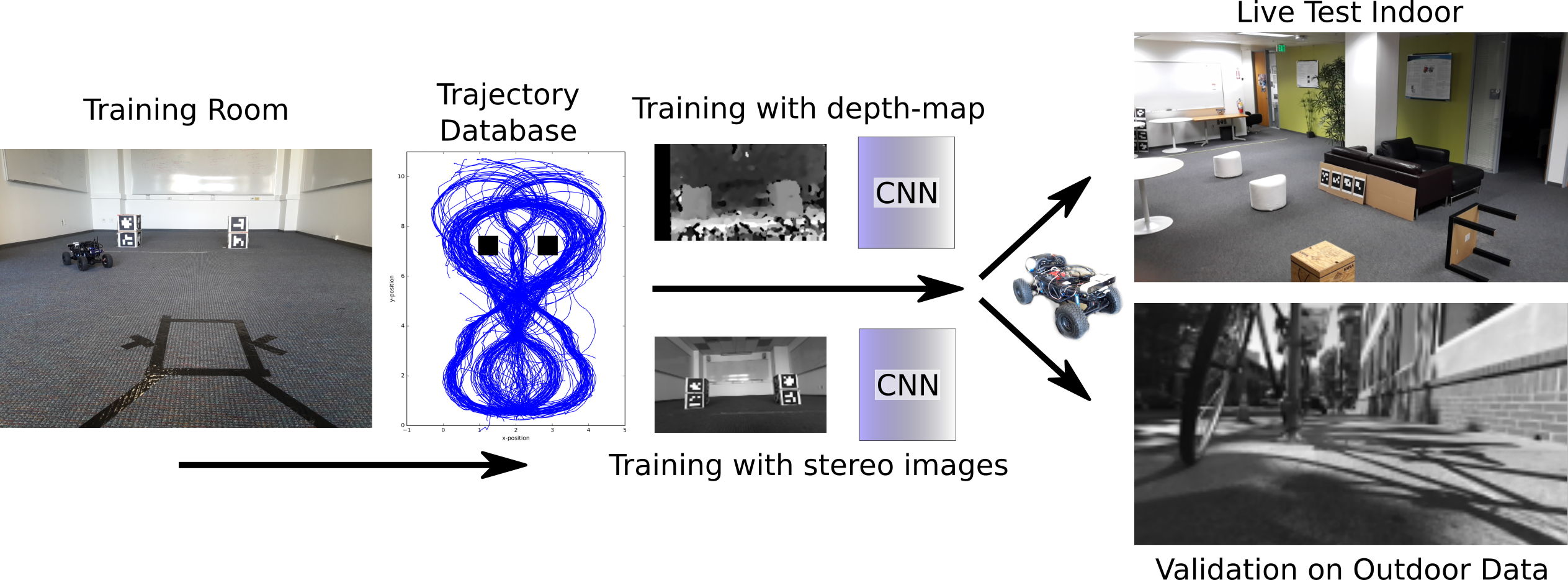}
    \caption{Overview of the method: Data from training is used to train the same network in two different ways. The driving, based on a depth-map and based on stereo images is compared in a cluttered room which the network was not trained in. }
    \label{fig:overview}
\end{figure*}  

A bigger problem with human behaviour cloning for autonomous driving is that there is no differentiation of all the human driving data. Good or bad human driving behaviours, regardless of drivers' intentions for driving in particular ways, are regarded as golden standards to be imitated. Data collection is not only labor intensive, but also lacking consistency, flexibility, and generalizability.

We propose an efficient and better alternative that simply puts our model car inside a small room, practicing automatic free roaming according to an optimal path planner based on depth images. It practices reaching a random target location by executing the driving action supplied by a path planner. It examines the free space measured by the depth image from camera inputs at the moment. Its planner calculates the optimal trajectory and decides on optimal steering signals. Such trials can go on continuously with human intervention, yielding (depth, steering) data for a deep net to learn from. 

Our method has important advantages. With the path planner, we have fine control over trajectories, e.g. we can set and control appropriate safety distances, or drive the shortest path when several are possible. With a universal spatial representation such as depth images, we can easily transfer the driving behavior to environments with unseen and different visual appearances. Depth images can be derived on current embedded platforms as the leveraged NVIDIA Jetson-TX1 in real-time, robustly and accurate from stereo image pairs \cite{Hirschmuller2005}.   

Our experimental results demonstrate that, after mere hours of practicing in a small room with two simple obstacles, our model car can roam a very cluttered environment. More excitingly, our model car can also function well in a never seen outdoor environment even without any adaptation.

The paper is structured as follows, After summarizing and placing our method in the general research domain, we discuss the most related work in autonomous driving and behavioral cloning in \cref{sec:related_work}. In \cref{sec:method} we present our model car framework, give specifics on our path planner and network training regime. We also give examples of depth images used in training and perceived during test time. In \cref{sec:experiment} we describe the performed experiments in our test environment and show all driven trajectories from a top-view. Finally we show a comparison of the performance of training and testing in between stereo and depth-image based training indoors and on our outdoor dataset. In supplementary released video material the test in a cluttered office room is shown. We are committed to publish the source-code for training and image conversion along with the paper.

%% file: 3related.tex
\section{Related Work}
 \label{sec:related_work}
{\bf Autonomous Driving.}
Comparable work on Behavioral Cloning of steering and motor signals for model cars shows that it is possible to train a shallow convolutional neural net (CNN) to learn navigation in various terrains, in snow conditions and even at night \cite{Hou2017}. Many recent autonomous driving approaches use deep neural networks to solve various tasks, from sensor processing and scene interpretation to action selection and control \cite{Huval2015,Wulfmeier2017,Wulfmeier2017a}. \cite{Codevilla2017} and \cite{Chowdhuri2017} use external meta-information at a deeper encoding level of the network to select a specific desired variation of the driving behavior, after training. Their methods allows to choose a certain path at an intersection or drive more at the center or at the side of a path when with pure behavioral cloning several options would be equally probable.

{\bf Path planner based driving.}
\cite{Pfeiffer2017} uses a path planner in a simulation to navigate to a number of goals with several obstacles. They apply their algorithm in the real world to navigate based on 2D-laser range findings. An external goal position and extracted features from a CNN are fused into their final fully connected layers to produce steering and motor commands towards that goal. A motion planner is used as expert to train the network though this is performed in a deterministic simulation. 
We see advantages in real world training opposed to \cite{Pfeiffer2017}: The state progression of a model car while driving is probabilistic, allowing for a natural exploration of the state space. It is not necessary to add artificial noise to the control or input signal as when wanting to explore a larger state-space as compared to training in simulation. Using a path planner in this real-world setting still minimizes the need for human intervention. In order to even improve state-space exploration the path planner can be used to track the network task execution and intervene only to avert a collision, similar to the technique used in \cite{Bojarski2016a}.

\cite{Gupta2017} designed a learning architecture for a robot in a simulation, which builds a cognitive abstract map and plans actions in that map at the same time. The map is built and updated from vision and knowledge about the robot's own motion. In their work they exploit the fact that a turn of the own robot will lead to a known rotation in their derived map and consequently they rotate their belief about the world depending on the egomotion. Successful navigation and also semantic task performance, as finding an office chair, could be shown in simulation though the application is limited by the assumption of perfect odometry.

{\bf Depth from Images.}
End-to-end learning from vision to actions was successfully used in driving tasks \cite{Bojarski2016a} and robotic manipulation \cite{Levine2015}. In concurrent work, \cite{Tai2017} train a mobile robot to evade pedestrians in hallways, based on raw depth input, which they use to generalize examples from simulation to the real world. They employ a generative adversarial imitation learning strategy (GAIL) to learn temporal correlations in their training data. In contrast, in our approach we present ten subsequent frames concatenated to the network to learn temporal correlations, as is further explained in \cref{sec:network_training}. In future work employing GAIL in our setup could prove a valuable addition. 

In the robotics domain \cite{Tai2016a} and \cite{Tai2016b} were some of the first to perform behavioral cloning and reinforcement learning on raw depth input, in their case from a Microsoft Kinect camera. In the same year \cite{Bojarski2016a} drove a car down an offroad path using behavioral cloning based on stereo vision instead. In both approaches, human input is needed to build the dataset for the CNN to train on. \cite{Hernandez-Aceituno2016} showed examples where object detection for autonomous navigation using a Kinect camera outperforms two older stereovision approaches at speeds lower than $15 km/h$. However, even though dedicated depth sensors could be used for our method, we focused on retrieving depth from stereo images as this can be scaled more easily to full-sized cars operating at higher speeds. Recent advances in depth prediction even from monocular images reinforce our assumption that depth data for navigation is available on a variety of systems.\cite{Godard2016,Kuznietsov2017,Zhou2017}.

%% file: 4method.tex
\section{Method}
\label{sec:method}

%We drive automatically in a training room to create exhaustive driving examples. 
%learn general navigational capabilities 
%
%We use a depth map during training to generalize to other environments ...
%
%We show that we are successful by experiments in a different room ...
%
%At the end we show how the approach generalizes to the outside domain...

In a first stage the dataset of stereo-vision images, reconstructed depth-images and steering commands from many driven trajectories is generated using the path planner. With the acquired data we train two different networks with almost the same design, apart from the size of the input layer. \Cref{fig:overview} shows an overview of the method and our cluttered test-room. Next we describe the model-car framework and path planner used.

%\end{multicols}
\begin{figure*}[htpb]
  \centering
    \begin{subfigure}[b]{0.48\textwidth}
    \centering
    \includegraphics[width=0.74\textwidth]{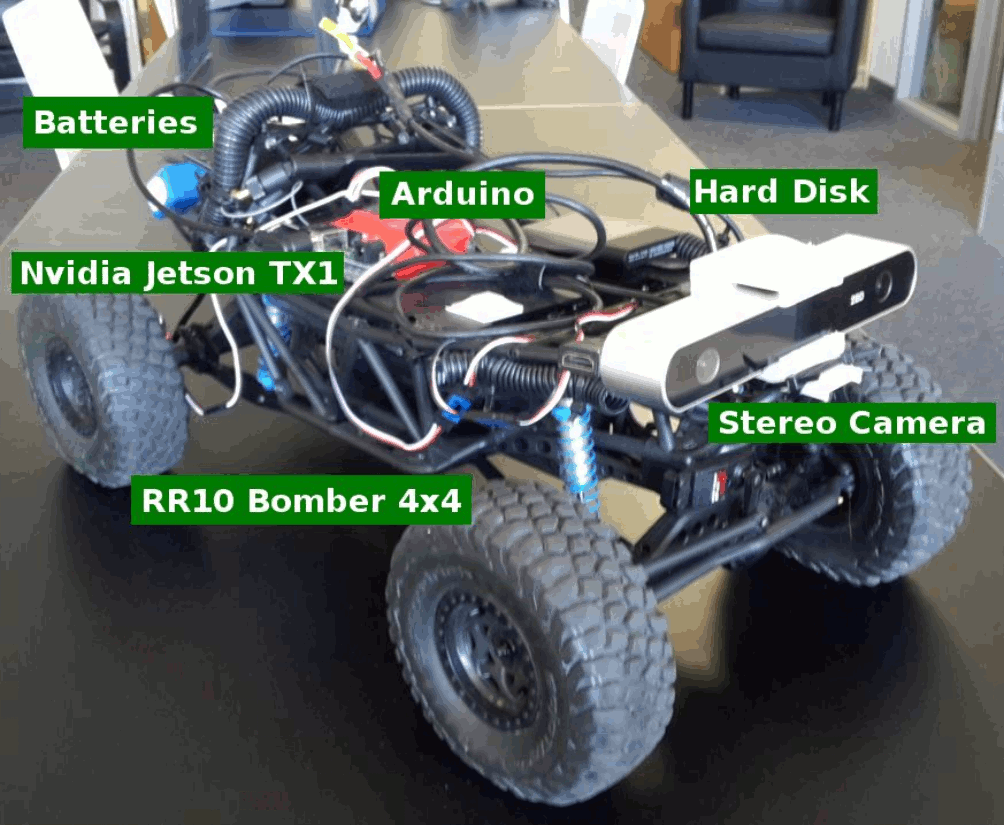}
    \caption{Model car components}
    \label{fig:car_with_labels}
    \end{subfigure}
    \hfill
    \begin{subfigure}[b]{0.48\textwidth}
    \centering
    \includegraphics[width=0.9\textwidth]{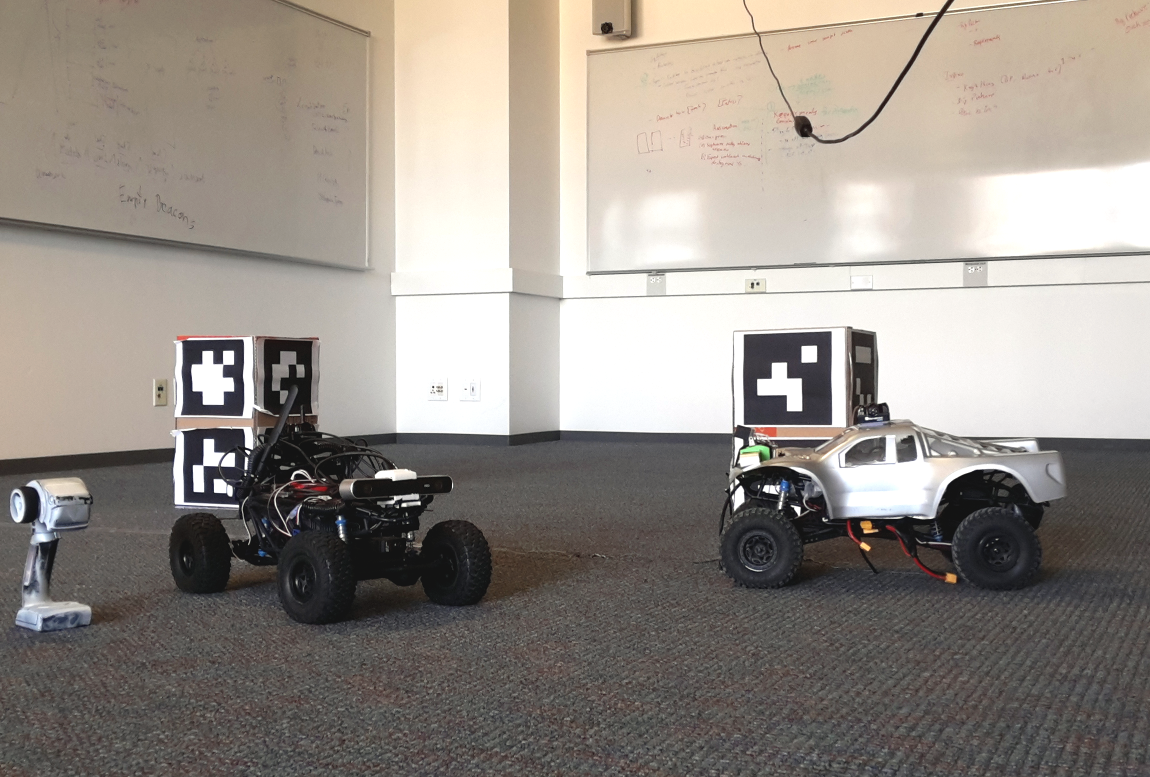}
    \caption{Two model cars with controller in front of two obstacles}
    \label{bots}
    \end{subfigure}
    \caption{The model car platform we built consists of many cars, though only one was used in the experiments described here.}
    \label{fig:model_car_platform}
\end{figure*}

%\begin{multicols}{2}

\subsection{Model Car Framework}

A testbed of several autonomous model cars has been built. \Cref{bots} show two model cars in the training room. \cref{fig:model_car_platform} shows the model car components: An NVIDIA Jetson TX1 with embedded GPU built on a RR10 Bomber 4x4 RC car. Available sensors are a ZED Stereo Camera from Stereolabs at the front, a gyroscope and a wheel encoder. The turning radius is approximately $1.2~m$ which is also set at the path planner.

\FloatBarrier
\subsection{Path Planner}
\begin{figure*}[!ht]
    \centering
    \includegraphics[width=0.7\textwidth]{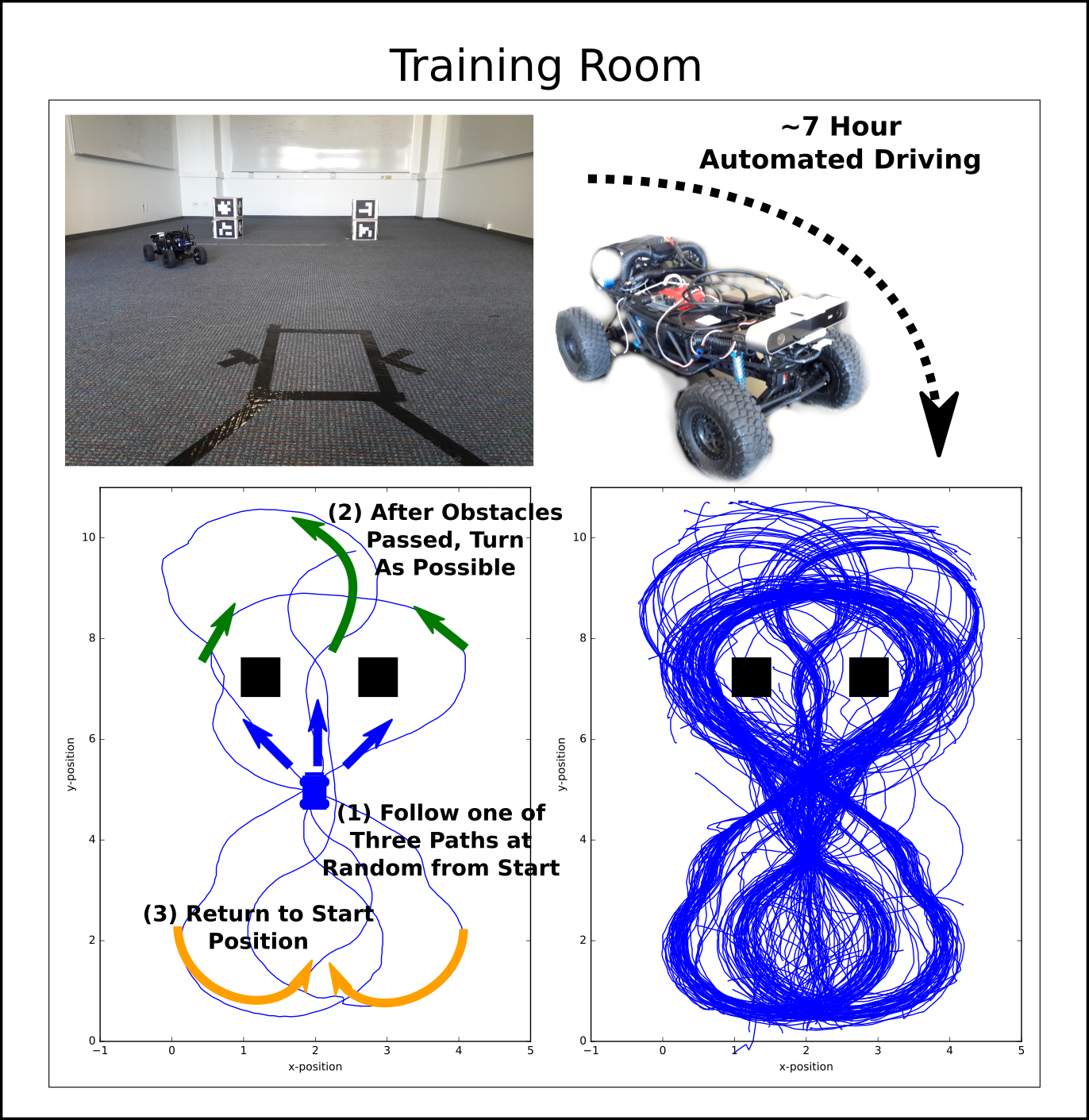}
    \caption{Training procedure in the training room. The car starts at the start position (1) and the path planner drives it at random to one of three positions next to the obstacles, shown as black squares. At the end of the room (2) it manoeuvres to drive back along one side of the room, depending on the available space. Finally it turns at the far end (3) and returns to the start position to begin a new trajectory, always with some variance depending on the pose when arriving at the start. After approximately 7 hours of driving a high coverage of the room with trajectories is reached. Shown at the bottom right in blue are 200 trajectories from a total of 1003. They are not shown because they cover the entire space quite dense, outlined by the 200.}
    \label{fig:training_room}
\end{figure*}

We let a Dubins-model based path planner drive on randomized trajectories based on ground truth x,y coordinates and a fixed map. The trajectories can be tuned to set e.g. turning radius, desired distance to obstacles and the boundaries of the map to match the manoeuvrability of the model car. Each trajectory is considered to start at the center position where the path planner calculates a feasible path to one of three positions: left of, right of and in the center in between of two obstacles, as seen in  \cref{fig:training_room}. Once one of these positions is reached there are further positions along the room which are planned depending on the chosen position to lead the car along the back of the room back to the starting position. When the car drives in between the two obstacles, a complex turning behavior has to be performed, going first in one direction and then into the other, to overcome the limitations of the turning radius. The exact trajectory towards the planned path is tracked with a PID-controller and additional correction via cross-track error. With our method, data recording was possible with minimal human intervention and the model car had to be retrieved from a stuck state only when the localization system failed to report the correct location, leading to a collision.

\FloatBarrier
\subsection{Network Training}
\label{sec:network_training}

\begin{figure*}[!h]
  \centering
    \includegraphics[width=1.0\textwidth]{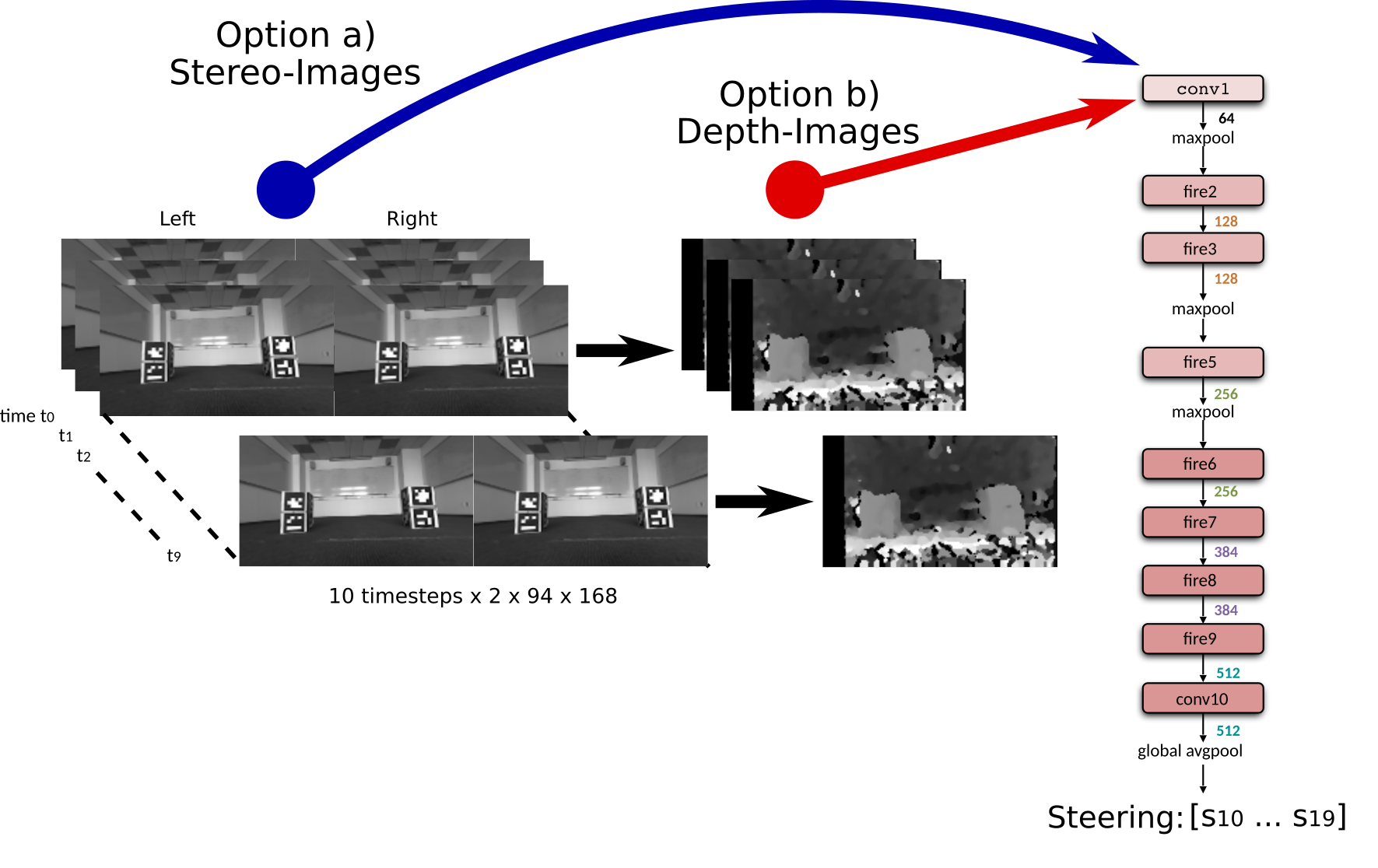}
    \caption{We compare two approaches, option a) which we employed successfully in the past and option b) which reaches the generalization results described in this paper. In both cases the Squeezenet-based network \protect\cite{Iandola2017} receives at the first layer images from 10 timesteps, concatenated along the channel dimension. Fire layers extract features and compress information further along the channel dimension. At the end the network predicts 10 future steering commands.}
    \label{squeeze}
\end{figure*}

The used network architecture is based upon SqueezeNet as developed by \cite{Iandola2017}, a network designed for image classification which performs well on our embedded platform. We take temporal correlation over frames into account by concatenating frames over 10 time-steps. Input images are scaled down to a resolution of $94 \times 168$ for a single input image. The input to the network, as shown in \cref{squeeze}, is therefore either $3\times 2\times 10\times 94\times 168$ in case of stereo color-images or $10\times 94\times 168$ in the case of concatenated gray-scale depth-images. At the last layer, a vector of steering and motor commands over 10 time-steps is predicted by regression using a 2d convolution with 10 kernels. Only the steering command, closest in time will be used for steering, though the other are generated as a side-task. This is motivated by results in the field which indicate that training a specific task improves when multiple related tasks have to be solved at the same time \cite{Caruana1995,Seltzer2013,Chowdhuri2017}. It favours the prediction of entire trajectories over single points of control through the car.

%\clearpage

The network was trained to predict the actions of the path planner over 16 epochs. Images in the dataset were recorded at 30 fps and the output steering commands are compared against ground-truth values, 33~ms apart. The speed is fixed and not part of the network's prediction. Steering commands are compared against the ground truth with a Mean Squared Error-loss. Learning is performed with an adaptive learning rate method, which is described in \cite{Zeiler2012}.%, implemented in PyTorch.

\subsection{Depth-Map from Stereo Images}

The motivation in presenting depth images to the network is to increase in-variance against differences in appearance of the input scene. They are created with the SGM stereo method (\cite{Hirschmuller2005}, blocksize 5, number of disparities 16). We selected these parameters through hand-tuning, trying to achieve a compromise in between high details in the distance and less noise, while we tolerated remaining errors in depth-reconstruction. It is visible that errors in depth-reconstruction, e.g. from a colored patch at the ground, are moving consistent with that patch for a few frames. This leads to black and white shapes floating through the visual field, depending on the movement of the camera. We hypothesize, though could not yet further investigate, that this supports network training through a richer optical flow. The released supplementary video material contains scenes showing reconstructed depth images to illustrate the phenomena.

\begin{figure*}[h]
    \centering
    \begin{subfigure}[t]{0.48\textwidth}
        \centering
        \includegraphics[width=\textwidth]{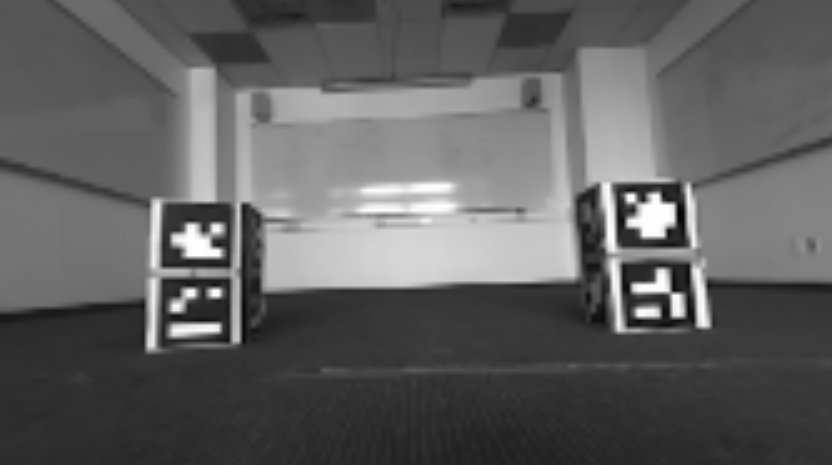}
        \caption{Original perceived images of room with two obstacles}
        \label{fig:aruco}
    \end{subfigure}
    \hfill
    \begin{subfigure}[t]{0.48\textwidth}
        \centering
        \includegraphics[width=\textwidth]{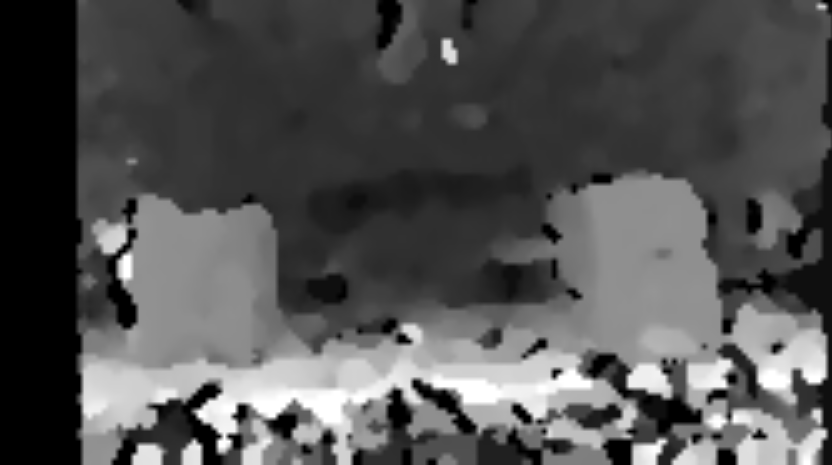}
        \caption{Depth reconstruction. Obstacles begin to disappear}
        \label{fig:aruco_depth}
    \end{subfigure}
    \centering
    \begin{subfigure}[t]{0.48\textwidth}
        \centering
        \includegraphics[width=\textwidth]{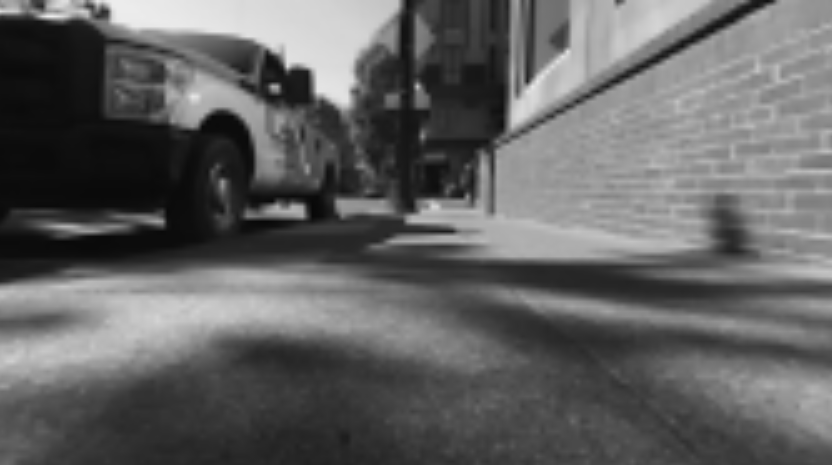}
        \caption{Driving on a sidewalk}
        \label{fig:sidewalk}
    \end{subfigure}
    \hfill
    \begin{subfigure}[t]{0.48\textwidth}
        \centering
        \includegraphics[width=\textwidth]{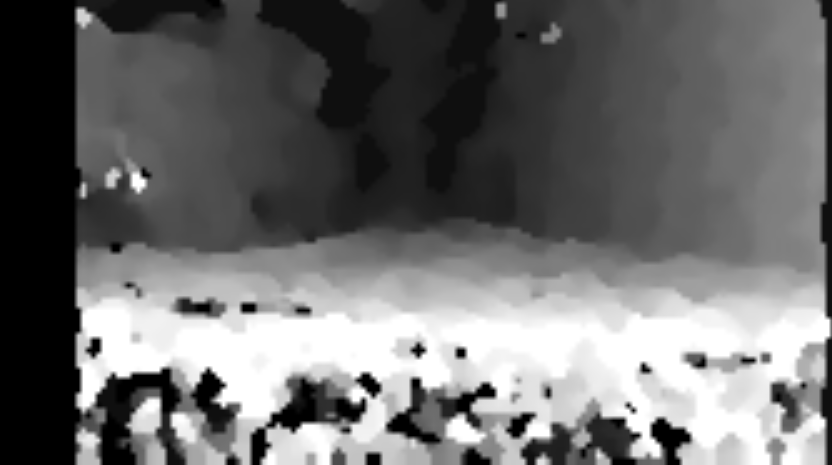}
        \caption{Depth reconstruction of the sidewalk}
        \label{fig:sidewalk_depth}
    \end{subfigure}
    \caption{Top row: The training environment indoors shown as raw camera image next to the reconstructed depth image (\subref{fig:aruco} and \subref{fig:aruco_depth}). Bottom row: Raw camera image next to depth image outdoor on a sidewalk (\subref{fig:sidewalk} and \subref{fig:sidewalk_depth})}
    \label{fig:aruco_sidewalk}
\end{figure*}

\begin{figure*}[h]
    \centering
    \begin{subfigure}[t]{0.48\textwidth}
        \centering
        \includegraphics[width=\textwidth]{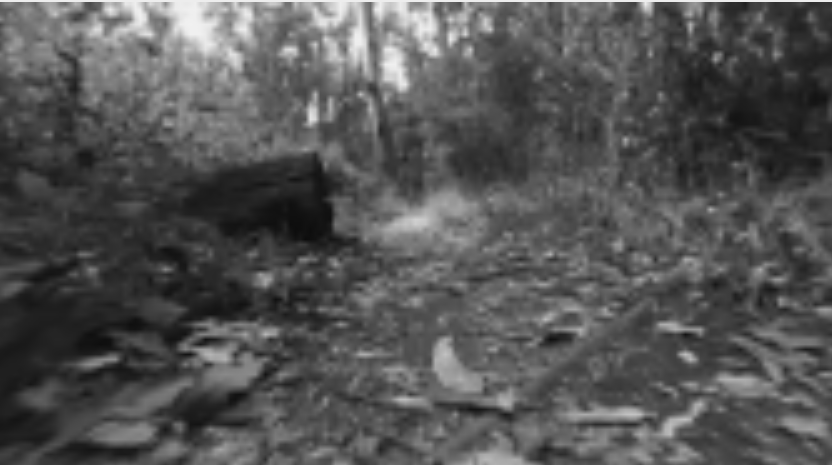}
        \caption{A path in a nearby park}
        \label{fig:woodpath}
    \end{subfigure}
    \hfill
    \begin{subfigure}[t]{0.48\textwidth}
        \centering
        \includegraphics[width=\textwidth]{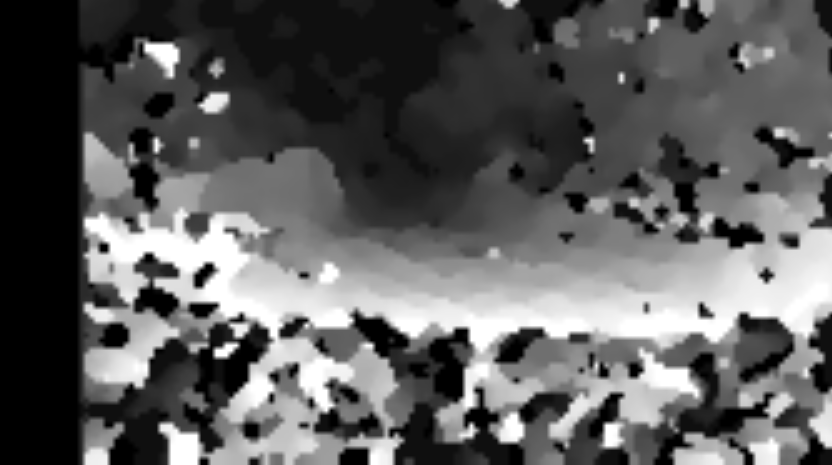}
        \caption{Heavy noise though still a visible gradient along the path}
        \label{fig:woodpath_depth}
    \end{subfigure}
    %\caption{More images. Eventually delete}
    %\label{fig:more_images}
%\end{figure}
%\begin{figure}[h!]
    \centering
    \begin{subfigure}[t]{0.48\textwidth}
        \centering
        \includegraphics[width=\textwidth]{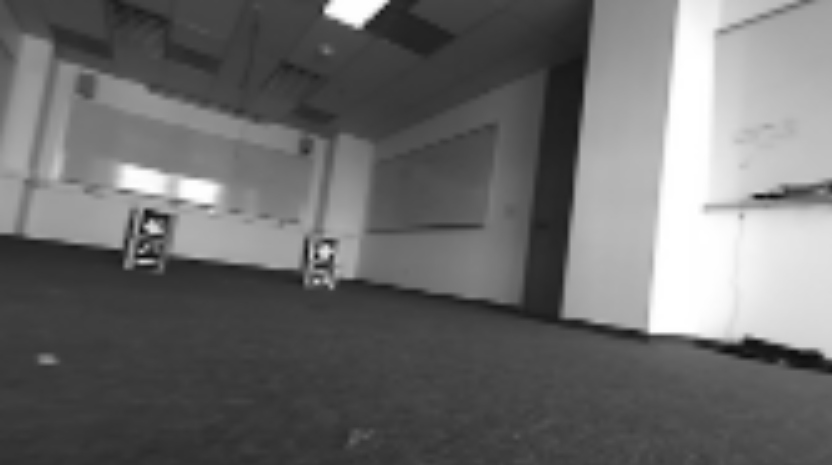}
        \caption{Training room with two obstacles, further away}
        \label{fig:room_again}
    \end{subfigure}
    \hfill
    \begin{subfigure}[t]{0.48\textwidth}
        \centering
        \includegraphics[width=\textwidth]{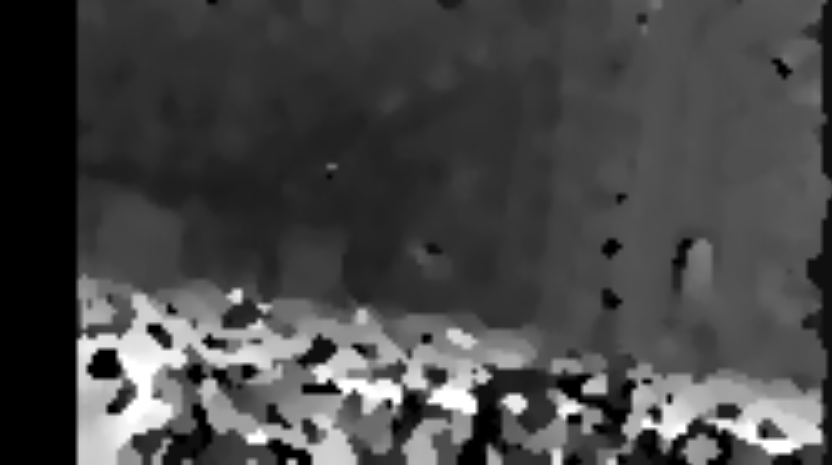}
        \caption{Depth reconstruction of the left image. Reconstruction errors are visible at the bottom}
        \label{fig:room_again_depth}
    \end{subfigure}
    \caption{Top row: One of the original stereo images from driving in a park (\subref{fig:woodpath}) with the reconstructed depth image (\subref{fig:woodpath_depth}). Bottom row: The training room again for comparison (\subref{fig:room_again}) and the depth image (\subref{fig:room_again_depth})}
    \label{fig:more_images2}
\end{figure*}

The lower difference in appearance can be seen in \cref{fig:aruco_sidewalk}, where the obstacles in the training room are shown next to their depth image. Walls appear as continuous gradients while obstacles are gray and change their brightness close to the camera. The same gradient on walls can be seen in an example, where we drove the cars outside on sidewalks, seen in \cref{fig:sidewalk} and \subref{fig:sidewalk_depth}. In \cref{fig:more_images2} the comparison of depth images from recordings on a path in the woods and the training room can be seen. Even though more fragments are present, the path is visible as gradient towards the center of the image.

% \newpage

%% file: 5results.tex
\section{Experimental Evaluation}
\label{sec:experiment}
\begin{figure*}[h]
    \centering
    \begin{subfigure}[t]{0.48\textwidth}
        \centering
        \includegraphics[width=\textwidth]{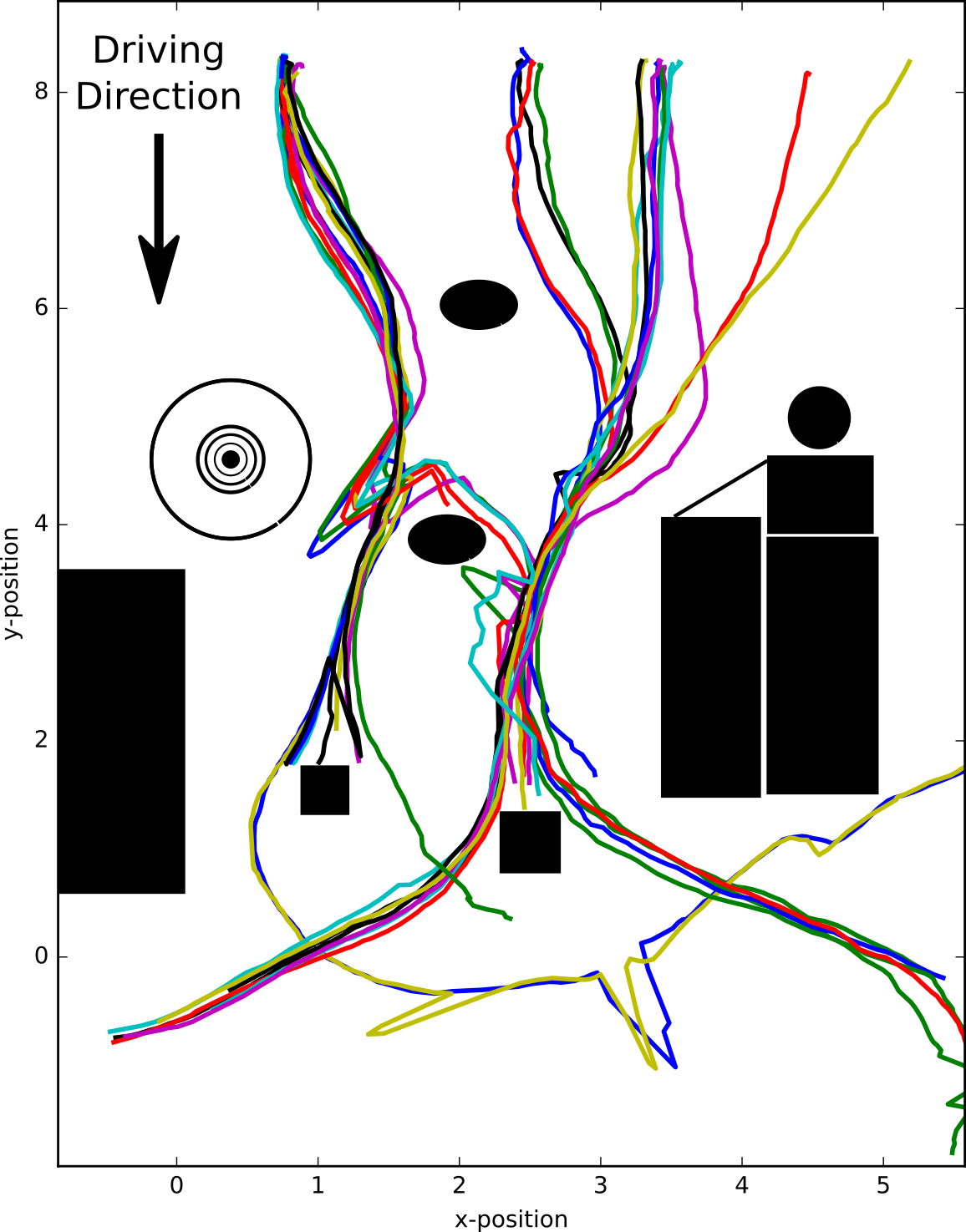}
        \caption{Trajectories driven with depth-image trained model in novel environment.}
        \label{fig:depth_trajs}
    \end{subfigure}
    \hfill
    \begin{subfigure}[t]{0.48\textwidth}
        \centering
        \includegraphics[width=\textwidth]{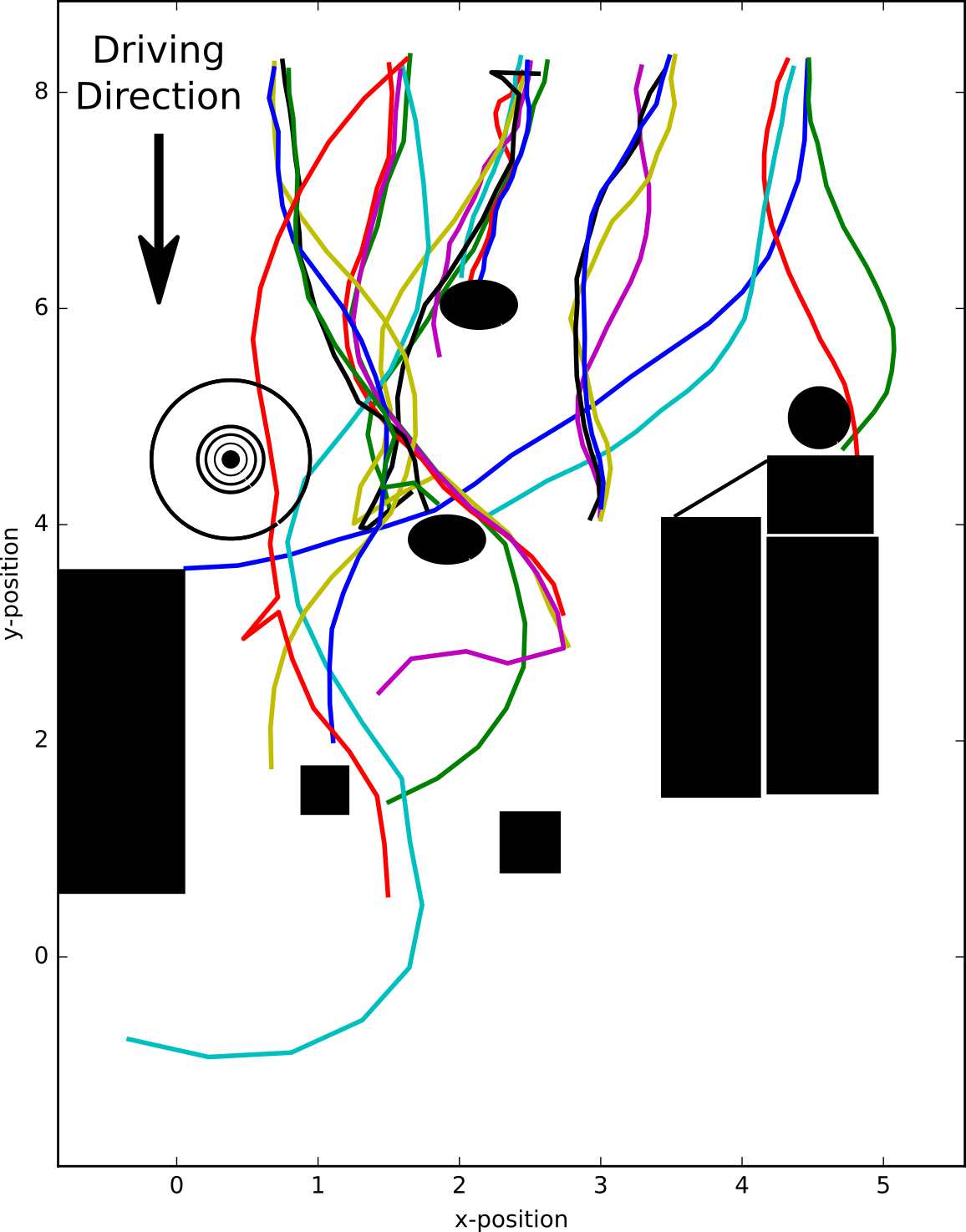}
        \caption{Trajectories driven with stereo-image trained model in novel environment}
        \label{fig:stereo_trajs}
    \end{subfigure}
    \caption{Comparison of trajectories in test-environment. Obstacles are shown as black forms, one table with a smaller based and slim table-leg is shown as a number of circles. Some errors of the localization system are visible as short spikes, which over the length of the trajectories are considered negligible. Driving direction is from the top to the bottom.}
    \label{fig:traj_comp}
\end{figure*}

Performance across domains is tested by measuring obstacle avoidance in a novel environment. A testing-room with a more complicated layout and more and different obstacles is traversed by the model car in several trials from a range of starting positions. Each trial is ended once the car reaches the other side of the room or drives into one of the obstacles. Driving behavior from the network, trained only on raw depth data is compared with the network, trained only on stereo video data. In the end the average length of the trajectories is compared. \Cref{fig:stereo_trajs} shows higher amounts of early ended trajectories, showing a collision or the car being stuck after passing an obstacle too close. After a preliminary study showed best results, the network was trained for only one epoch before testing. Even though the validation value still improves in later epochs we hypothesize this indicates an over-fitting effect to the training-data early in training, which is not apparent from comparing training and validation loss. This could indicate that our validation data, which was collected in the same room as the training data, is too similar to the training data. In the future experiments with different network iterations across epochs, trained with a lower learning rate, can help to optimize the driving behavior further.
% \FloatBarrier

\begin{figure*}[h!]
    \centering
    \includegraphics[width=0.7\textwidth]{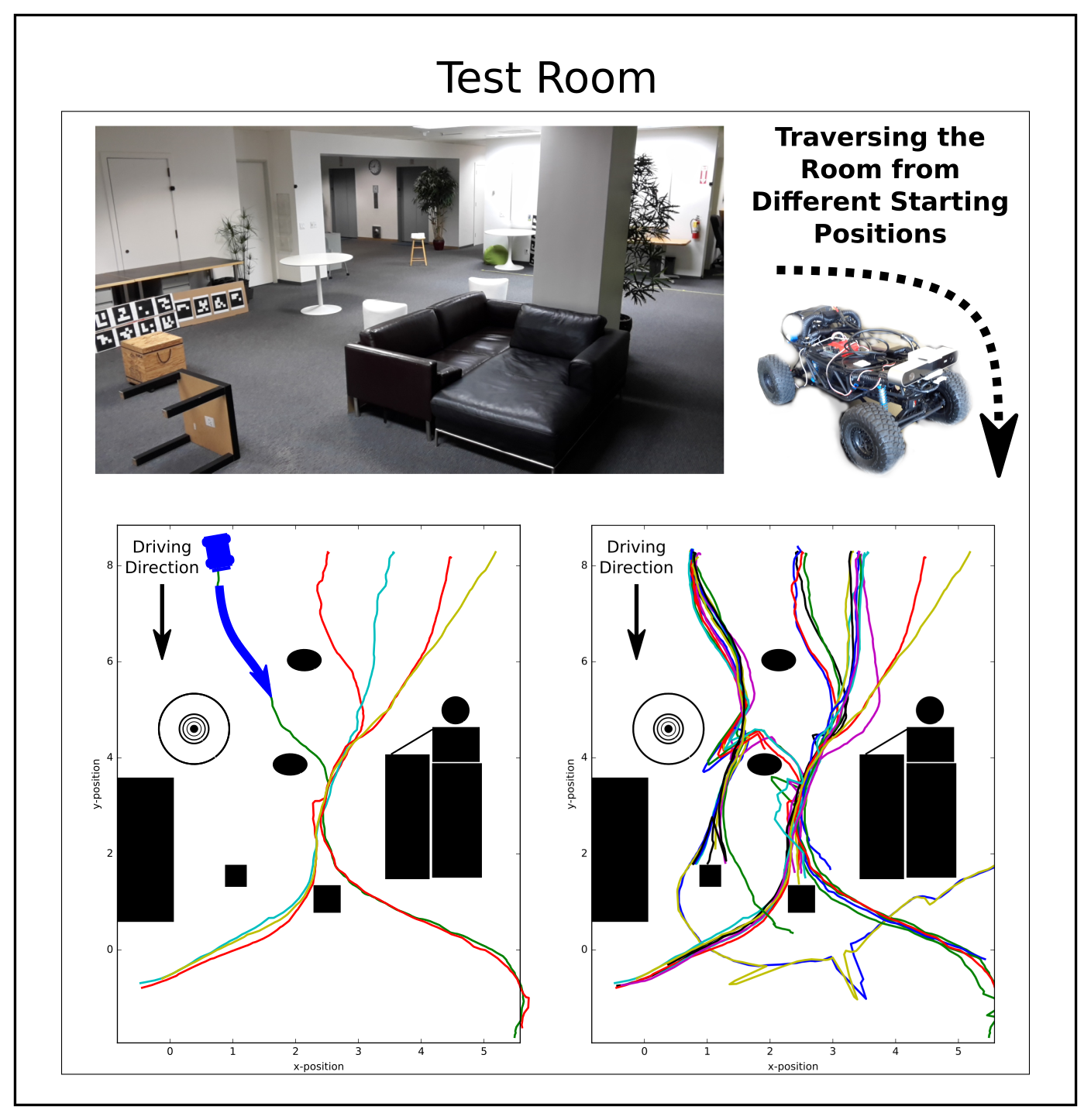}
    \caption{Driving from different starting positions in the test room. The model car starts driving at the top facing the other side of the room. No explicit goal is given apart from driving straight when that is possible. All trajectories are recorded with a localization system.}
    \label{fig:test_room}
\end{figure*}    

\begin{figure*}[h!]
    \centering
    \includegraphics[width=0.7\textwidth]{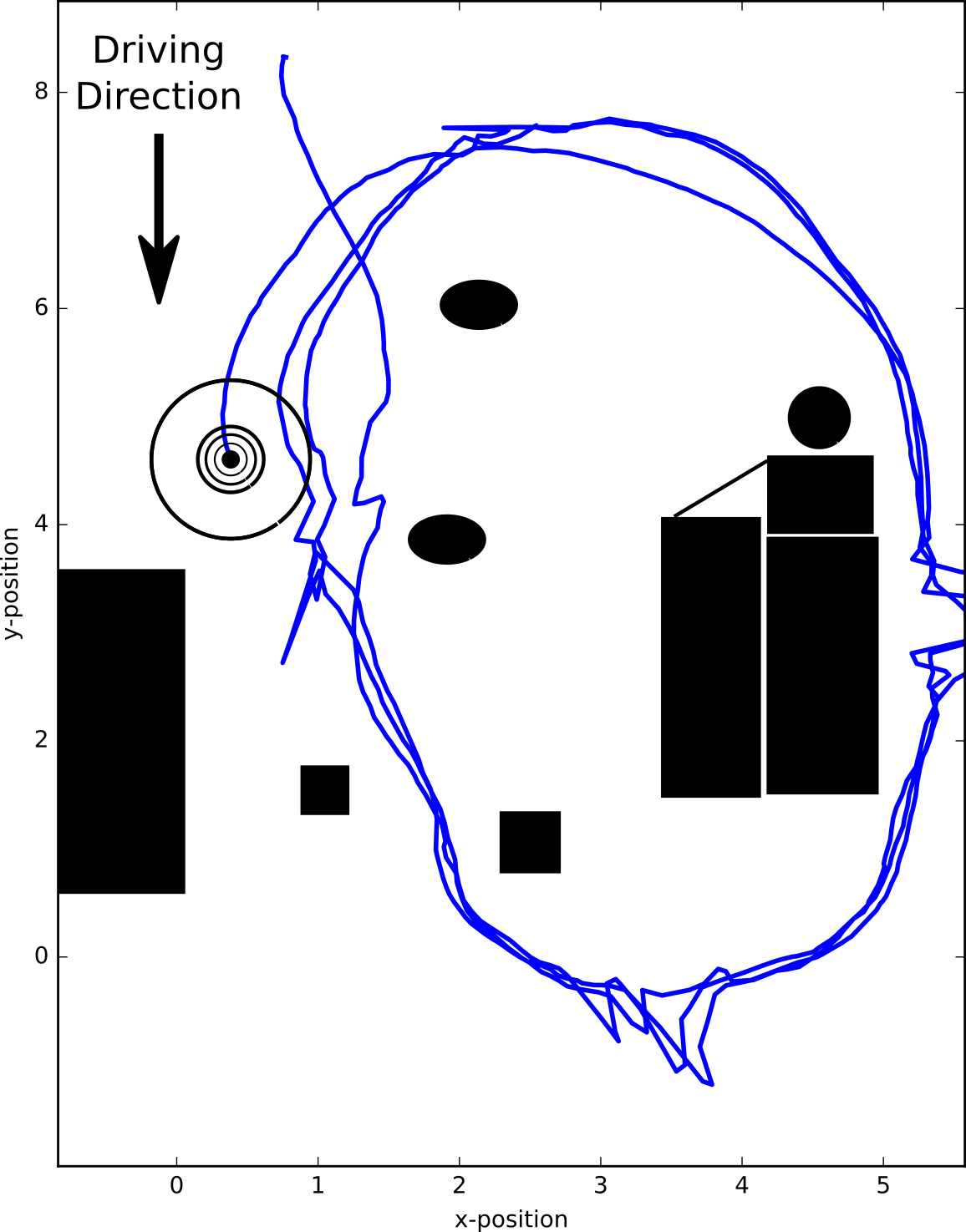}
    \caption{An outlier case where the car was able to roam through the room for three complete rounds until coming to a halt. It emphasizes good wall avoidance, especially in the narrow passage to the right.}
    \label{fig:outlier}
\end{figure*}

The test environment seen in \cref{fig:test_room} is significantly more cluttered than the training space seen in \cref{fig:aruco}, \cref{fig:room_again} and \cref{bots}. However, in \cref{fig:outlier} it can be seen that in one trial the vehicle was able to combine obstacle and wall avoidance to a degree that several turns led to continuous roaming. As this was not part of the test-task we consider the trial as an outlier which was removed from the results so the increased length would not bias the comparison of trajectories. In  \cref{fig:depth_trajs} and \cref{fig:stereo_trajs} the driven trajectories are shown. Apart from the improved average driven trajectory using the depth-images, it can be qualitatively seen that failure cases are more systematic. 

\setlength{\tabcolsep}{4pt}
\begin{table}[h!]
\begin{center}
\caption{Results of depth-image against stereo vision-based driving, compared by their achieved trajectory length with given standard deviation $\sigma$}
\label{table:measurings}
\begin{tabular}{l|p{0.15\columnwidth}|p{0.15\columnwidth}|r|r}
\hline\noalign{\smallskip}
Network & $\#$ Trajectories Driven & Avg. Length & $\sigma$ Length &Longest Trajectory \\
\noalign{\smallskip}
\hline
\noalign{\smallskip}
Stereo Trained  & 24 &  $5.32~m$ & $2.22~m$ & $11.23~m$ \\
Depth Trained & 28 &  $\textbf{9.78}~m$ & $3.09~m$  & $\textbf{18.80}~m$ \\
\hline
\end{tabular}
\end{center}
\end{table}
\setlength{\tabcolsep}{1.4pt}

Even though figure \cref{fig:stereo_trajs} shows some successful collision avoidance manoeuvres, the number of collisions is higher, as can be also seen in the average driven distance, shown in \cref{table:measurings}. Furthermore, the longest trajectory driven by the depth-image trained model is about $60\%$ longer. 

The failure cases in \cref{fig:depth_trajs} (trajectories ending at an obstacle) indicate classes of objects and incoming angles under which collisions can be reproduced. We hypothesize that this stems from the inability to favour one evasion path over another in head-on scenarios where both evasion actions are equally probable. This is a known problem with Behavioral Cloning and can be improved in the future by using the meta-information, described in \cref{sec:related_work} to randomly choose on direction. 

% \setlength{\tabcolsep}{4pt}
% \begin{table}[!h]
% \begin{center}
% \caption{Results of depth-map against stereo vision-based driving}
% \label{table:measurings}
% \begin{tabular}{lr}
% \hline\noalign{\smallskip}
% Stereo-Image Driving & \\
% \noalign{\smallskip}
% \hline
% \noalign{\smallskip}
% Average Length of Trajectories  & $5.32~m$ \\
% Standard Deviation & $2.22~m$\\
% Amount of Trajectories Driven & 24 \\
% Longest Trajectory Driven & $11.23~m$\\
% \hline\noalign{\smallskip}
% Depth-Image Driving & \\
% \noalign{\smallskip}
% \hline
% \noalign{\smallskip}
% Average Length of Trajectories  & $\textbf{9.78}~m$ \\
% Standard Deviation & $3.09~m$\\
% Amount of Trajectories Driven & 28 \\
% Longest Trajectory Driven & $\textbf{18.80}~m$\\
% \hline
% \end{tabular}
% \end{center}
% \end{table}
% \setlength{\tabcolsep}{1.4pt}

\subsection{Comparison with outdoor data}

\begin{figure*}[h!]
    \centering
    \begin{subfigure}[t]{0.48\textwidth}
        \centering
        \includegraphics[width=\textwidth]{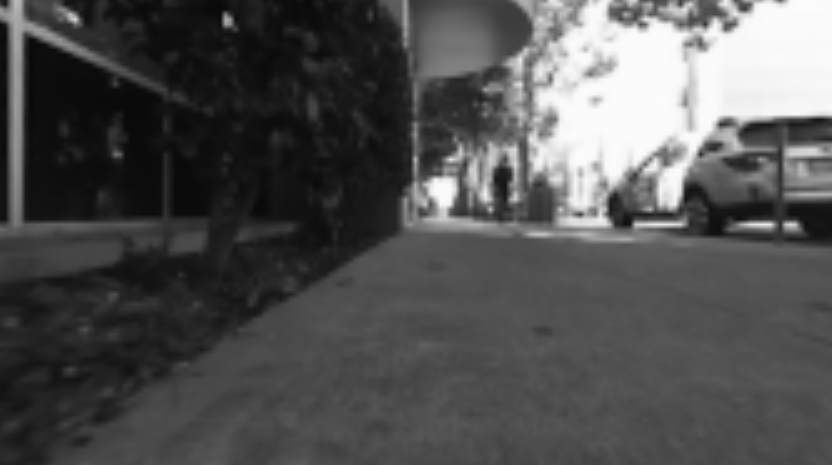}
        %\caption{Trajectories driven with depth-image trained model in novel environment.}
        %\label{fig:depth_trajs}
    \end{subfigure}
    \hfill
    \begin{subfigure}[t]{0.48\textwidth}
        \centering
        \includegraphics[width=\textwidth]{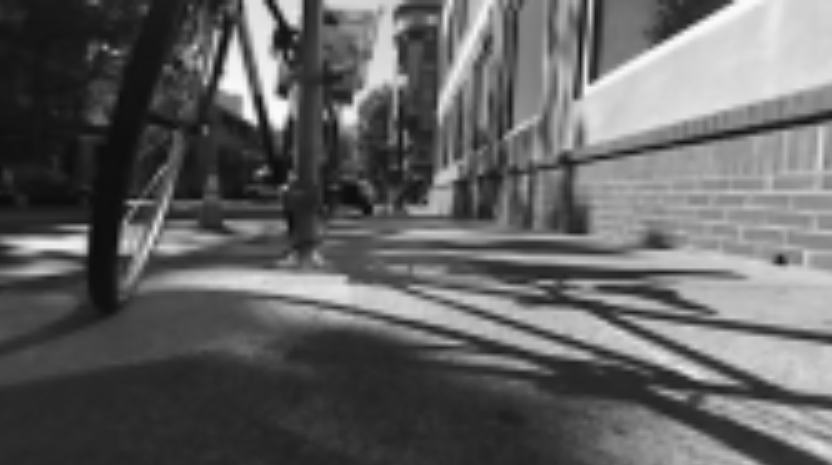}
        %\caption{Trajectories driven with stereo-image trained model in novel environment}
        %\label{fig:stereo_trajs}
    \end{subfigure}
    \caption{Images of sidewalk-driving which is a large part of the outdoor dataset}
    \label{fig:outdoor}
\end{figure*}

In addition to the cluttered office space we tested the indoor-trained models on previously recorded outdoor data. Two representative scenes of driving on sidewalk are shown in \cref{fig:outdoor}. While in the office space no label for the test data exists, outdoor steering and motor labels from a human drivers are recorded. 

\begin{figure*}[h!]
    \centering

    \begin{subfigure}[t]{0.49\textwidth}
        \centering
        \includegraphics[width=\textwidth]{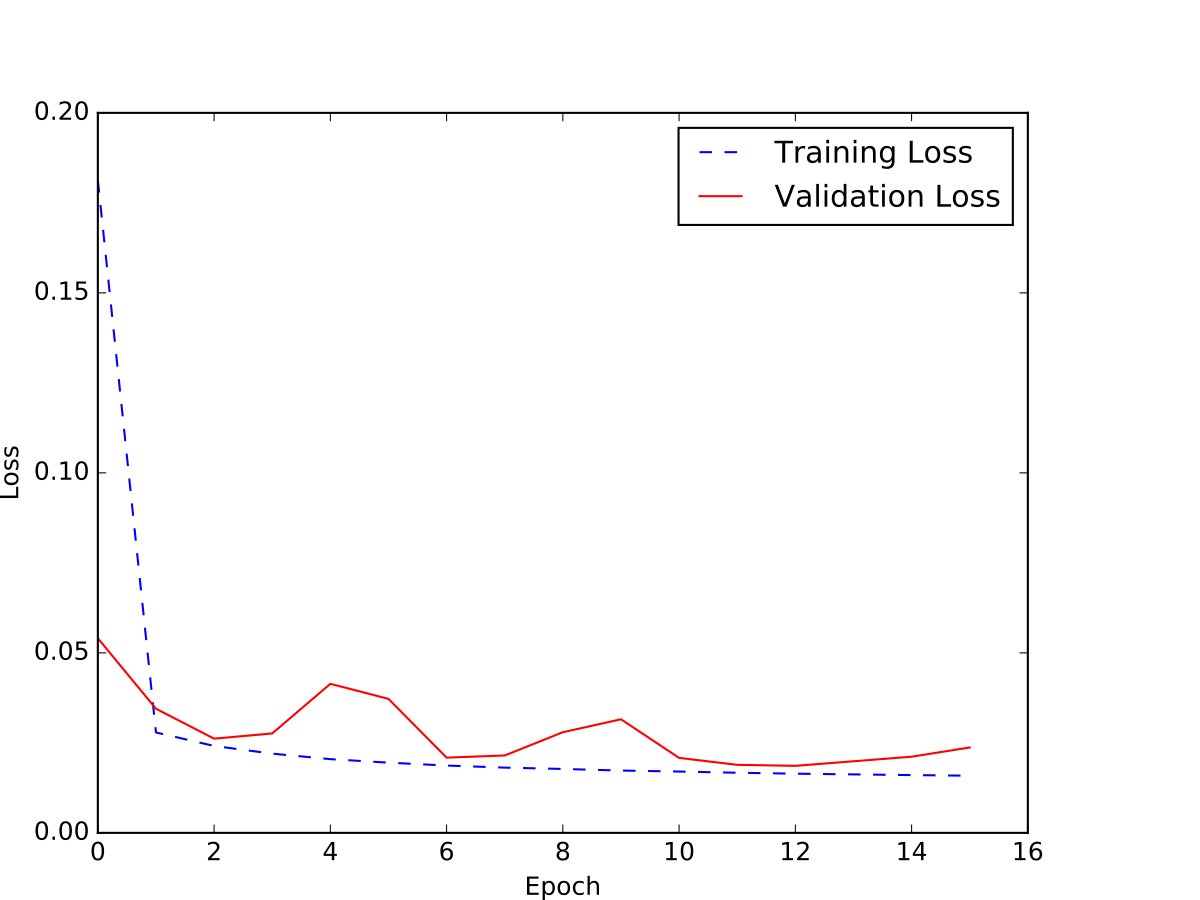}
        \caption{Training results when training with depth images}
        \label{fig:training_performance}
    \end{subfigure}\hfill
        \begin{subfigure}[t]{0.49\textwidth}
        \centering
        \includegraphics[width=\textwidth]{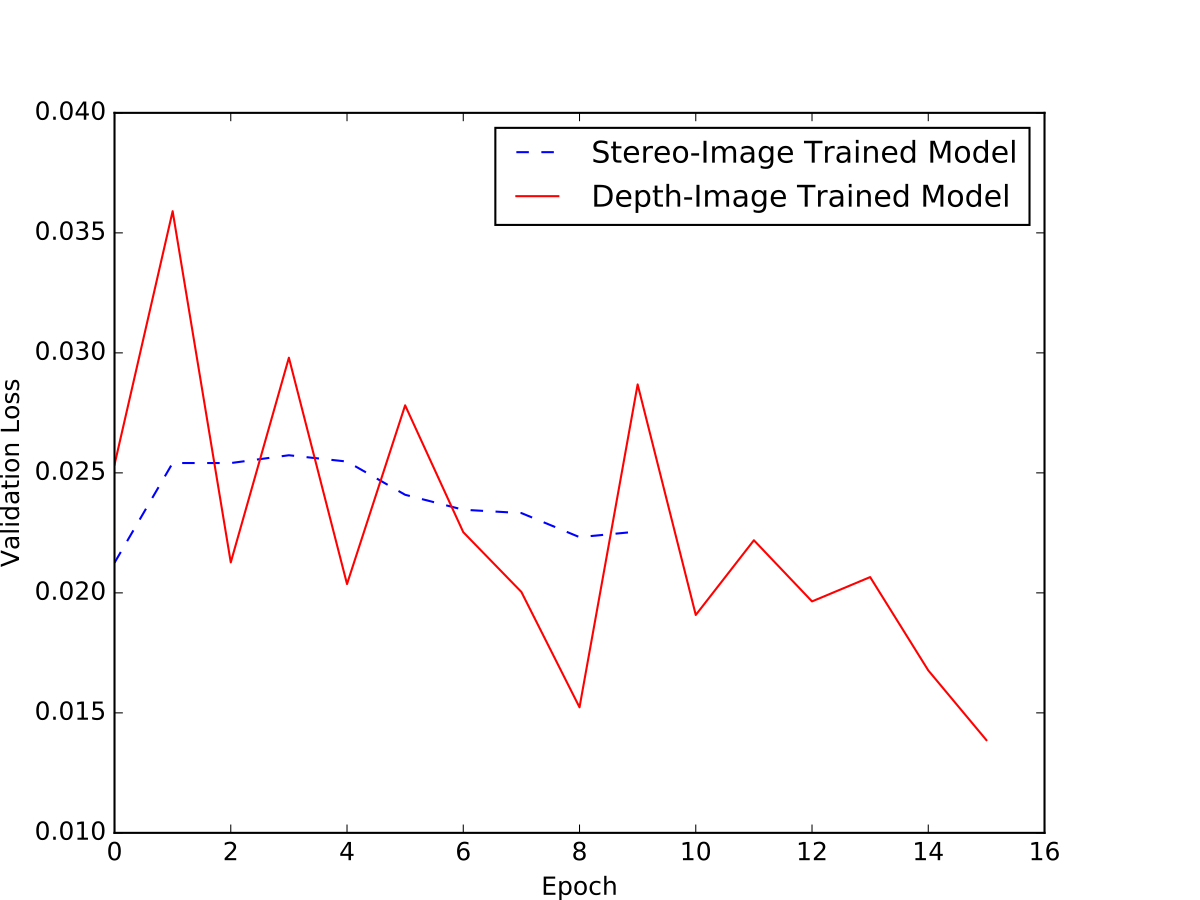}
        \caption{Outdoor data-validation, comparing steering angle regression with labeled human steering decisions.}
        \label{fig:outdoor_test}
    \end{subfigure}
    \caption{Shown are loss and validation loss as Mean Squared Error. The indoor training in (\cref{fig:training_performance}) shows good convergence during training and robustness against over- and under-fitting. Tests of different epochs of training on outdoor data is shown in figure (\subref{fig:outdoor_test}). The depth-image trained network varies more in its performance over epochs than the stereo-image trained network, however good converging performance can be seen after 14 epochs. Training on stereo-image data showed slowly converging performance, showing in general good generalization to outdoor driving with both models. The stereo-trained model was trained over 9 epochs as of the time of this writing though the performance of the depth-trained model is shown over all 16 epochs.}
    \label{pic:test_env}
\end{figure*}

We compared the performance of the two trained networks with the human performance, driving on sidewalks, even though the interpretation of the results is not straightforward. The network is trained to predict the output of a path planner, configured to drive optimal trajectories around obstacles. The human in our dataset can perform according to different motivations as staying on one side of the sidewalk, anticipating pedestrians from an elevated viewpoint or avoiding chaotically moving obstacles as dogs opposed to predictable obstacles, as left furniture. Nevertheless, we can see in \cref{fig:outdoor_test} that both networks improve their approximation of the human behavior in later epochs, even though those are trained exclusively on indoor data. When comparing \cref{fig:training_performance} and \cref{fig:outdoor_test},  this leads to the surprising result that as the network improves its prediction of the path planner's desired trajectories, it also improves the prediction of human driving behavior. 

%% file: 6summary.tex
\section{Conclusion and Future Work}
\label{sec:conclusion}
We were able to train a network from scratch to navigate a model car in an unseen cluttered office space based on raw depth input using Behavioral Cloning. First we used a path planner for driving in a simple environment, with very sparse visual features. Our method minimizes human intervention and explores the real driving state-space while creating expert examples. During database creation, steering decisions are chosen based on optimal trajectories around obstacles, with variance due to the inherent probabilistic nature of real driving. We showed successful training of a network and generalization of driving behavior to a significantly more complex office room with many obstacles. We also compared the superior generalization properties of using depth images as input during training and driving. With depth-image based generalization, we were even able to show good steering angle prediction on an outdoor dataset, comparing the network's output, trained on path-planner generated data only, with steering from a human driver.

Future research is needed on how to include external hierarchical control into the free roaming method. Ambiguous situations as intersections or head-on scenarios with obstacles need to be resolved, e.g. with a hierarchical decision modules which would chose among equally probable actions. Furthermore, different metrics for testing can be developed by calculating the optimal paths in the cluttered room and comparing the network performance during training not only to separated data in the training room but to theoretically optimal trajectories, depending on the fixed testing environment. 
%\newpage